\documentclass[a4paper,fleqn]{cas-dc}

\usepackage[authoryear,longnamesfirst]{natbib}

\usepackage{multirow}
\usepackage{cuted}
\usepackage{caption}
\usepackage{graphicx}

\usepackage{tabularx}
\setcitestyle{notesep={; }}                  

\def\tsc#1{\csdef{#1}{\textsc{\lowercase{#1}}\xspace}}
\tsc{WGM}
\tsc{QE}

\begin{document}
\let\WriteBookmarks\relax
\def\floatpagepagefraction{1}
\def\textpagefraction{.001}

\shorttitle{}    

\shortauthors{}  

\title [mode = title]{Metamon-GS: Enhancing Representability with Variance-Guided Densification and Light Encoding}  



%

\author[1]{Junyan Su}
\fnmark[1]
\credit{Writing – Original Draft, Conceptualization, Validation, Software, Methodology, Investigation, Data curation, Visualization}


\author[1]{Baozhu Zhao}
\fnmark[1]
\credit{Writing - Review \& Editing, Supervision, Validation, Visualization}

\author[1]{Xiaohan Zhang}
\credit{Writing - Review \& Editing, Supervision, Visualization}

\author[1]{Qi Liu}[orcid=0000-0001-5378-6404]
\ead{drliuqi@scut.edu.cn}
\credit{Writing - Review \& Editing, Supervision, Resources, Project administration, Investigation, Funding acquisition}

\cormark[1]





\affiliation[1]{organization={Department of Future Technology, South China University of Technology},
            city={Guangzhou},
            postcode={511400}, 
            country={China}}







\fntext[1]{Equal Contribution}

\cortext[1]{Corresponding author}


\begin{abstract}
The introduction of 3D Gaussian Splatting (3DGS) has advanced novel view synthesis by utilizing Gaussians to represent scenes. Encoding Gaussian point features with anchor embeddings has significantly enhanced the performance of newer 3DGS variants. While significant advances have been made, it is still challenging to boost rendering performance. Feature embeddings have difficulty accurately representing colors from different perspectives under varying lighting conditions, which leads to a washed-out appearance. Another reason is the lack of a proper densification strategy that prevents Gaussian point growth in thinly initialized areas, resulting in blurriness and needle-shaped artifacts.
To address them, we propose Metamon-GS, from innovative viewpoints of variance-guided densification strategy and multi-level hash grid. The densification strategy guided by variance specifically targets Gaussians with high gradient variance in pixels and compensates for the importance of regions with extra Gaussians to improve reconstruction. The latter studies implicit global lighting conditions and accurately interprets color from different perspectives and feature embeddings.
Our thorough experiments on publicly available datasets show that Metamon-GS surpasses its baseline model and previous versions, delivering superior quality in rendering novel views. 
\end{abstract}


\begin{highlights}
\item We propose Metamon-GS to improve rendering performance in anchor-based model.
\item Improvement in the densification strategy can benefit the distribution of primitives.
\item Enhanced directional perception leads to stronger color representations.
\end{highlights}



\maketitle


\section{Introduction}
Advances in computer graphics and 3D vision have greatly improved the capability to create detailed 3D scenes from 2D images. This advancement is based on a long history of 3D reconstruction methods, such as Structure-from-Motion (SfM) \citep{sfm, ecsfm} and Multi-View Stereo (MVS) \citep{mvs,robustmvs, hybridmvs}. One of the notable advancements is 3D Gaussian Splatting (3DGS) introduced by \cite{3dgs}, which presents a different way to represent 3D scenes by utilizing elliptical Gaussian functions, also known as Gaussians. 3DGS extends the idea of representing 3D scenes using primitives similar to other point-based methods \citep{point, neuralpoint}, which consider Gaussian functions as primitives. A Gaussian point is defined by a set of learnable features, such as spherical harmonic components, scale, rotation, position, and opacity. This method allows for a smooth and continuously changing representation of the scene, making rendering more efficient and serving as a useful tool for reconstructing high-quality 3D scenes. Furthermore, a well-reconstructed point cloud can also support a variety of downstream tasks including keypoint extraction \citep{keypoint}, 3D object detection \citep{c2bgnet, ms23d, pfenet}, and point cloud encryption \citep{protect}.

        

Challenges still exist in enhancing the quality of reconstruction under specific conditions, despite the strides made in creating high-quality new view images. The main problem that reduces the quality of reconstruction is the failure to densify certain areas with sparse initial point clouds sufficiently. Not enough Gaussians can adequately represent these areas, causing the model to get stuck in local minima and resulting in blurred and needle-shaped artifacts.
The other challenge is that when light condition changes too sharply across different view directions, the appearance in these areas shows color degradation and loss of detail. 

To address these challenges, we propose Metamon-GS. Our approach involves using a variance-guided densification technique to pinpoint areas that need more Gaussians by analyzing the variance of color gradients. This technique identifies areas that have a high variation in color but a low gradient variation in position, effectively pinpointing Gaussians that need to be made denser. By emphasizing color differences rather than just position gradients, we can achieve better representation in areas that were previously not fully reconstructed, addressing the shortcomings of smoothing gradients across rendered pixels.

Furthermore, we also address the task of effectively interpreting color depending on various perspectives. Taking inspiration from Instant-NGP \citep{ingp}, our suggestion is to utilize a hash grid for encoding view-dependent features. We consider lighting conditions as a global attribute and incorporate directional information, initially stored in the anchor embeddings, into the hash grid. In the MLP input, the view direction vector is replaced with the hash grid encoding of the direction vector. This approach leads to more accurate view-dependent color decoding. 

We conducted extensive experiments on the Mip-NeRF 360 \citep{mipnerf360}, NeRF Synthetic \citep{synthetic}, and Tanks \& Temples \citep{tnt} datasets, aiming to showcase the advantages of our model over the baseline models. We also performed ablation studies to validate the effectiveness of our proposed methods. Here are our contributions:

\begin{itemize}

\item We propose to use a hash grid to encode lighting conditions, which enhances the quality of reconstruction in scenes with intricate lighting.
\item We propose a novel densification strategy guided by variance of pixel gradients to address problems arising from the gradient smoothing of rendered pixels. This method is primarily implemented with CUDA within the part of code of Gaussian rasterizer.
\item Experiments on various datasets show that our approach successfully tackles these challenges and outperforms the baseline model.
\end{itemize}

\section{Related Work}
\subsection{Neural Radiance Field}
Neural Radiance Fields (NeRF) represent a revolutionary technique that has demonstrated exceptional performance in novel view synthesis tasks \citep{synthetic}. NeRF employs Multi-Layer Perceptrons (MLPs), to implicitly represent 3D scenes. By estimating a radiance field and utilizing volumetric rendering \citep{drebin1988volume, levoy1990efficient}, NeRF can generate high-quality images from new viewpoints. 

In recent years, NeRF has achieved significant advancements in multi-scale anti-aliasing and efficient scene representations, further enhancing rendering quality and practicality. For multi-scale anti-aliasing, studies such as \citep{nerfpp, mipnerf, mipnerf360, barron2023zip} have introduce novel ray sampling strategies and multi-scale feature field construction to overcome certain artifacts. Meanwhile, works like \citep{tensorf, kplanes, nsvf, plenoxels} leverage tensor factorization and hierarchical sparse voxel structures to achieve improvements in training efficiency while preserving reconstruction fidelity. These advancements further advance NeRF models' rendering quality and practicality significantly. 

However, NeRF and its variants highly rely on ray marching and sampling, as well as the use of MLPs to estimate color and opacity, resulting in slow training and inference speeds. 
To address these issues, researchers have proposed various optimization methods \citep{mobilenerf, ingp, sun2022direct, Hedman_2021_ICCV} aiming to improve the training and inference efficiency while maintaining rendering quality.

\begin{figure*}[t]
\centering
\includegraphics[width=0.99\textwidth]{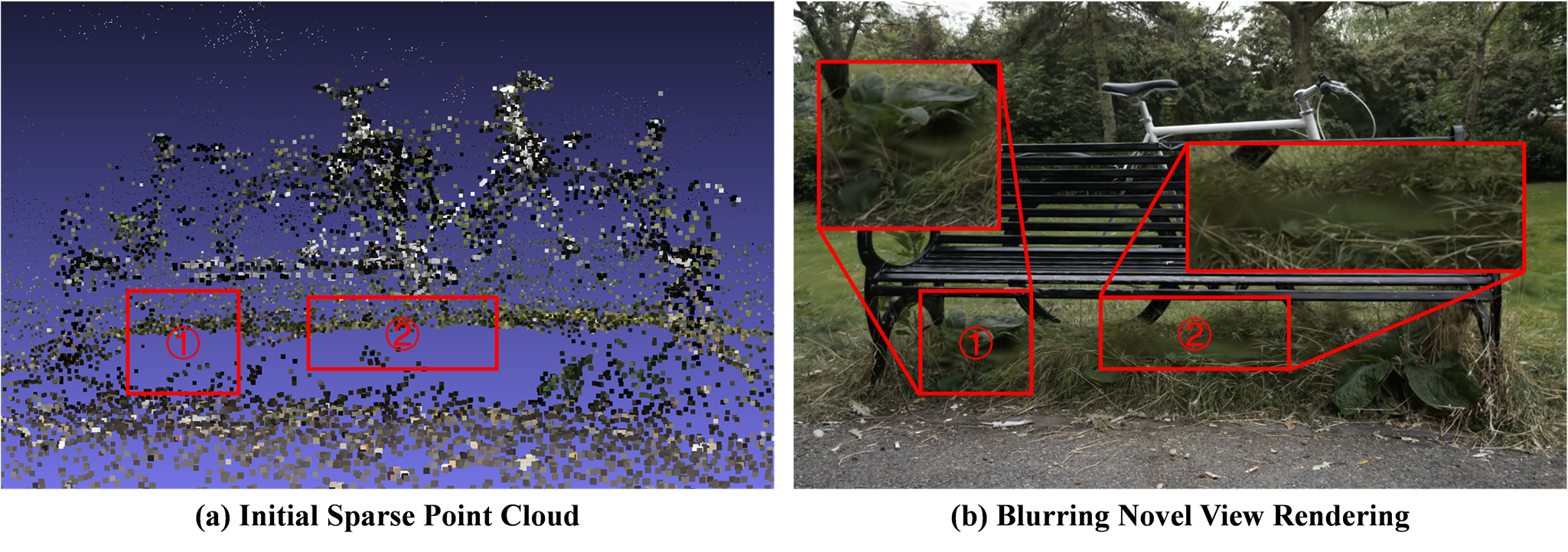} 
\caption{In certain areas, the Colmap-generated SfM point cloud is relatively sparse, as indicated by the red box in \textbf{(a)}. Utilizing this point cloud as the starting Gaussians and implementing the original clone and split density control strategy may lead to certain areas lacking enough Gaussians, ultimately causing an inadequate scene reconstruction, as demonstrated in the corresponding region highlighted by the red box in \textbf{(b)}. This greatly reduces the overall quality of novel view synthesis, especially in areas with intricate geometry or delicate details.}
\label{fig1}
\end{figure*}

\subsection{3D Gaussian Splatting and Variants}
Recently, the field of novel view synthesis has witnessed significant advancements, with 3DGS emerging as a particularly promising technique. Unlike traditional point-based methods \citep{differentiable, pointslam, lassner2021pulsar,kopanas2021point, insafutdinov2018unsupervised, yifan2019differentiable, dgpcfnerf}, 3DGS represents scene elements using ellipsoids defined by Gaussian functions, encapsulating both shape and color information. These ellipsoids, referred to as Gaussians, are characterized by a set of learnable features including spherical harmonic components, scale, rotation, position, and opacity. The 3DGS pipeline typically begins with the initialization of Gaussians, derived from sparse point clouds estimated by Structure-from-Motion (SfM) methods like COLMAP \citep{colmap}. During training, an adaptive densification mechanism is employed to split or clone these Gaussians, allowing for a more detailed scene representation. This process helps generate more Gaussians to model finer details in the scene, enhancing the quality of the synthesized views. As a result, 3DGS provides faster rendering speeds and produces higher-quality results in novel view synthesis.

Building upon the success of standard 3DGS, researchers have continued to further enhance its capabilities. One notable advancement was the development of anchor-based lightweight variants, such as Scaffold-GS \citep{scaffold} and Octree-GS \citep{octreegs}. For an anchor restricted inside a voxel, there are multiple offset Gaussians associated with it, with the features of individual Gaussians being embedded into one feature embedding. The densification of Gaussians in these methods is equivalent to the densification of anchors, which leverages the sparsity of local features of Gaussians, reducing the number of trainable parameters by decoding the feature embeddings of offset Gaussians with several MLPs.

\subsection{Densification Strategy}
For point-based methods, the initial point cloud often lacks sufficient points to fully model complex scenes, thus a densification strategy is needed to generate additional points. A probable situation is that the initial point cloud is imperfect, which presents the need for a proper densification strategy. 

3DGS \citep{3dgs} was designed with an adaptive densification strategy utilizing the gradient from the position of points in the NDC coordinate. This process involves splitting or cloning operations on target Gaussians, enabling more effective modeling of fine features and significantly improving the fidelity of reconstructed scenes. Effective though, this strategy still struggles to densify Gaussians in areas with intricate textures, where SfM methods produce insufficient initial points, as illustrated in Figure \ref{fig1}. Several studies have focused on improving densification strategies. For instance, FreGS \citep{fregs} enhanced the gradient magnitude in high-frequency areas by incorporating frequency domain supervision. This allows for more opportunities for Gaussians to be densified in areas with intricate details but may struggle with low-contrast regions. On the other hand, Pixel-GS \citep{pixelgs} improved upon the original approach by adjusting the average gradient based on the number of pixels rendered by each Gaussian point. This approach prioritizes larger Gaussians for densification, improving overall scene coverage, but potentially overlooking smaller, yet significant features.

Our Metamon-GS approaches the concept of reconstruction differently, using pixel color gradients for densification. The method we offer can be used in conjunction with current methods, potentially creating new possibilities for future research on adaptive densification techniques.

\subsection{View-dependent Color}
The appearance of a 3D scene can vary significantly depending on the viewing angle and lighting conditions, particularly due to surface properties like roughness. This phenomenon of view-dependent color poses a significant challenge in 3D scene reconstruction and rendering. Recent advancements in neural rendering and point-based reconstruction have made substantial progress in addressing this issue.

NeRF \citep{synthetic} incorporated camera orientation as an input to its MLP to generate view-dependent colors. This approach, while effective, can be computationally intensive and may struggle with large-scale scenes. 
Several Point-based methods \citep{3dgs, plenoxels, npbg, precomp} utilized spherical harmonics as point features to represent view-dependent colors efficiently. Spherical harmonics provide a compact representation of directional data, allowing these methods to capture color variations across viewing angles with relatively low computational overhead. 
Anchor-based variants of 3DGS such as Scaffold-GS \citep{scaffold} and OctreeGS \citep{octreegs} used feature embedding of an anchor instead of explicit features to describe Gaussians. In these methods, color is decoded by an MLP with feature embedding and view direction as input. Effective to some extent though, the use of feature embedding and MLPs for encoding view-dependent colors has shown limitations in capturing the complexity of lighting conditions.

To address this, our Metamon-GS develops a novel approach inspired by Instant-NGP \citep{ingp}, utilizing a hash grid for encoding view-dependent features. We treat lighting conditions, which are previously stored in anchor embeddings, as a global attribute and incorporate directional information into the hash grid. The view direction vector in the MLP input is substituted with the hash grid encoding of the direction vector within our approach. Our method improves view-dependent color decoding accuracy by incorporating lighting conditions into a hash grid.

\begin{figure*}[t]
\centering
\includegraphics[width=0.99\textwidth]{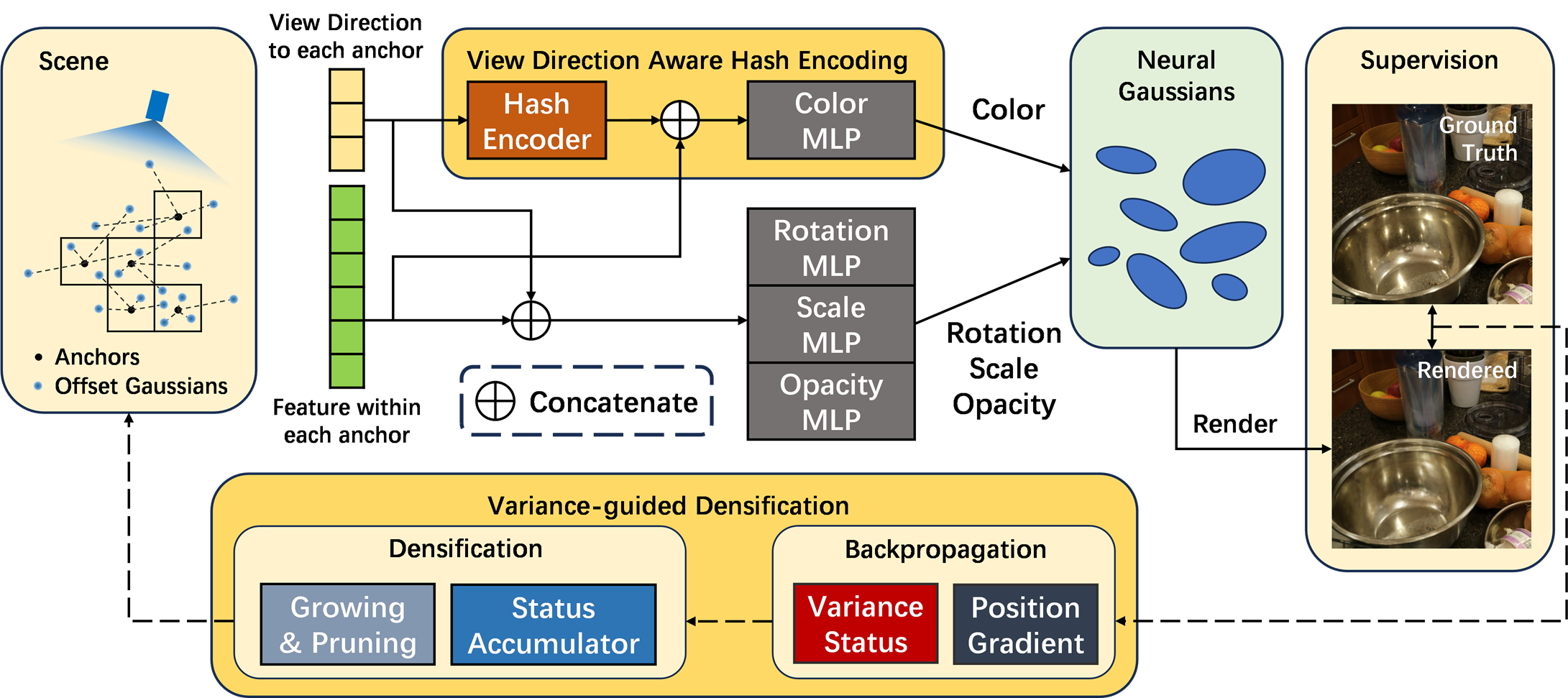}

\caption{\textbf{Overview of Metamon-GS.} Our method enhances 3DGS with two key innovations: (1) Integration of view-direction aware hash encoding to learn view-dependent features, e.g., lighting condition, and (2) A variance-guided densification strategy based on variance of color gradient during backpropagation. We first interpolate view-dependent feature embedding from a hash grid, concatenate them with anchor embeddings, and then feed them into the color MLP. Other features are decoded similarly to Scaffold-GS. This strategy, combined with our novel densification strategy, results in more accurate color representation and efficient Gaussians for scene representation.}
\label{fig4}
\end{figure*}

\section{Method}
Here, we propose Metamon-GS to address the aforementioned limitations, where Figure \ref{fig4} provides an overview of our approach. We first briefly review the original 3DGS densification strategy, conducting a pre-experiment to analyze the mean and variance of the color gradient of Gaussians. Then we introduce our Variance-Guided Densification strategy, explaining how it leverages the variance of color gradients. Finally, we present our Lighting Hash Encoder, which employs a hash grid to encode lighting information, enabling more accurate modeling of complex lighting conditions compared to original view direction inputs. 

\subsection{Densification in 3D Gaussian Splatting}
The original Gaussian densification strategy assumes that when a Gaussian point cannot fit the rendered pixels well, the point tends to shift its position due to a high backpropagated gradient on the Normalized Device Coordinates (NDC) position. The decision to densify the Gaussian point depends on the average gradient magnitude across visible viewpoints during the densification period. This strategy can be described as:
\begin{equation}
\begin{gathered}
    \bar{g}_\mathrm {norm} =\frac{\sum_{k=1}^{M}\sqrt{\left(\frac{\partial L_{k}}{\partial x_{ k}}\right)^{2}+\left(\frac{\partial L_{k}}{\partial y_{k}}\right)^{2}}}M \\
    \text{where}\ \frac{\partial L_{k}}{\partial x_{ k}}=\sum_{p}G_{p,k}
\end{gathered}
\end{equation}
where $M$ is the number of viewpoints, $G_{p,k}$ is the gradient from the $p$ th pixel the Gaussian $k$ rendered and $(x,y)$ is the NDC coordinate of the point. A Gaussian is selected to split and clone when its $\bar{g}_\mathrm {norm} $ exceeds the threshold $\tau_\mathrm {th}$.

This approach is effective in many cases, but it does not effectively concentrate when the backpropagated gradients from various pixels within the Gaussian's covered area evenly enhance its position in different directions. The resulting gradient from adding up these individual pixel gradients is quite small, preventing the Gaussian point from splitting. 

\begin{figure}[!h]
\centering
\includegraphics[width=0.47\textwidth]{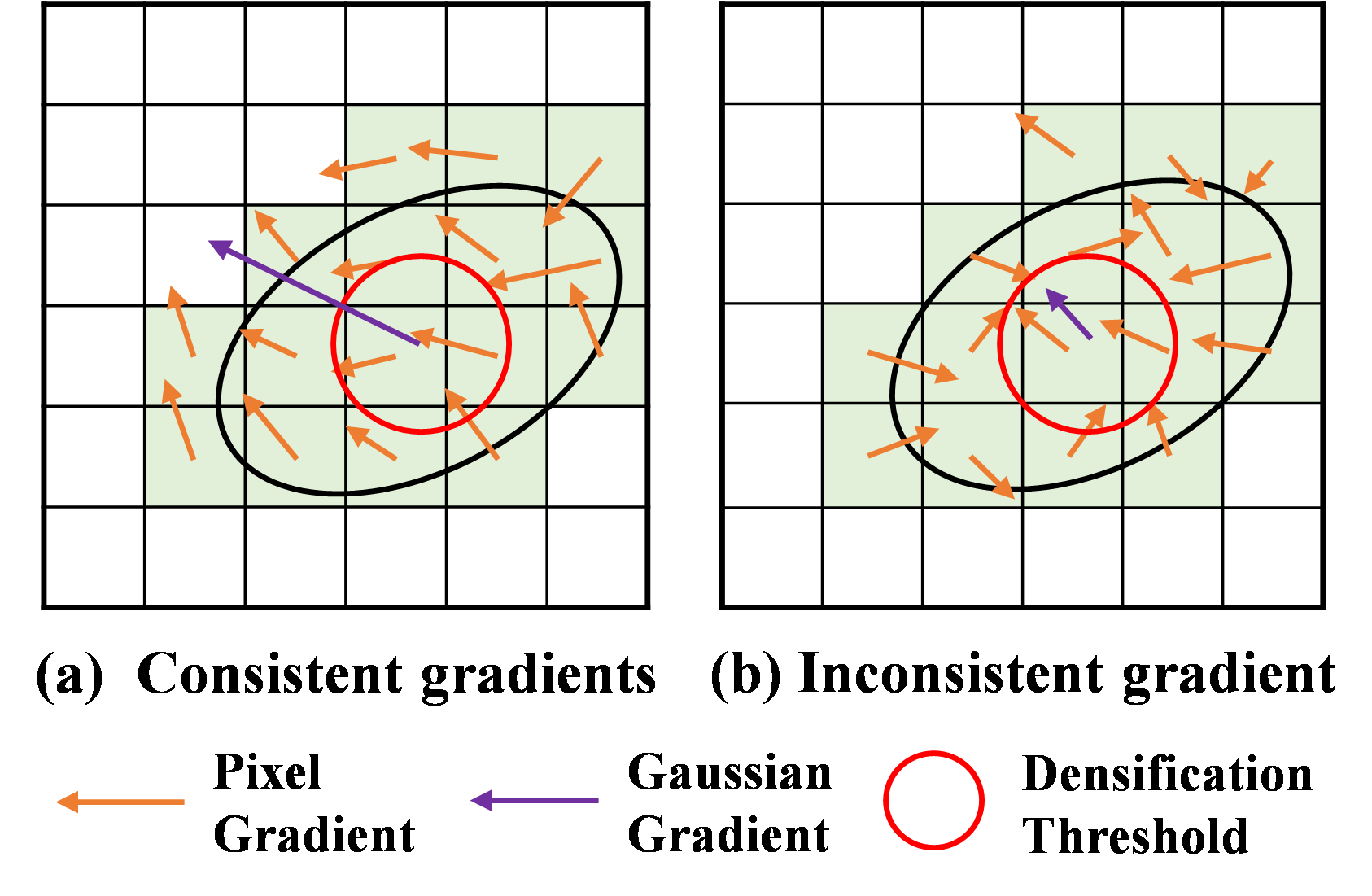} 
\caption{\textbf{Pixel-wise gradients and overall gradients of Gaussians under two different cases of under-reconstruction.} (a) In an ideal scenario, gradients of pixels consistently converge, forming a high Gaussian positional gradient, which allows effective densification. (b) In contrast, in a scenario with complex textures, gradients diverge in different directions, resulting in a lower Gaussian positional gradient, hindering proper densification.}
\label{fig2}
\end{figure}

We illustrate these two instances of under-reconstructed that occur during the implementation of adaptive Gaussian densification using positional gradient. In the first scenario, as depicted in Figure \ref{fig2}(a), the pixel gradients converge coherently, leading to a significant Gaussian positional gradient that facilitates densification. However, as illustrated in Figure \ref{fig2}(b), this assumption that under-reconstructed Gaussians exhibit high gradients fails to hold in certain scenarios. The gradients diverge, causing the Gaussian positions to be optimized in different directions. This is because the Gaussian covers a specific area with intricate textures. The divergence of gradients causes a decrease in the overall positional gradient following the weighted sum, preventing the Gaussian from reaching the necessary threshold for proper densification. The quality of rendering descends when there are not enough representative Gaussians.

\begin{figure}[h]
\centering
\includegraphics[width=0.99\columnwidth]{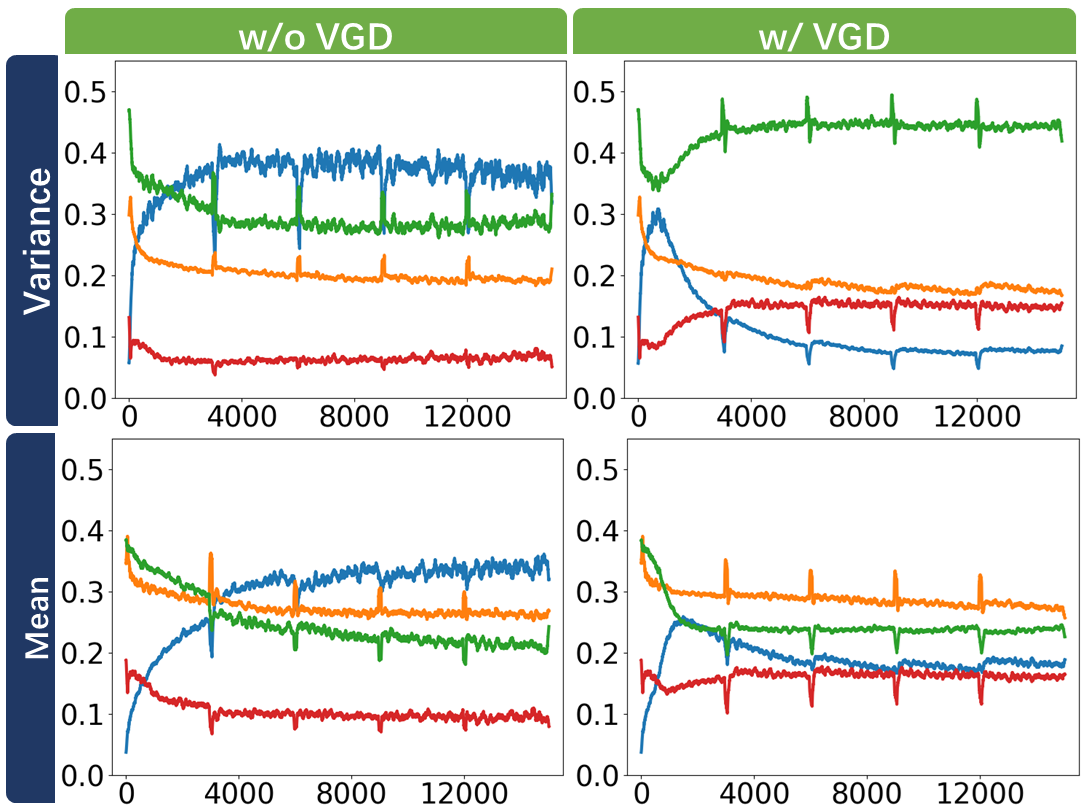} 
\caption{\textbf{Pre-experiment on 3D-GS with our proposed VGD strategy.} We employ an identical starting point for the reconstruction process. After 600 iterations, the curves for 3D-GS with VGD begin to disperse significantly from those without VGD. For 3D-GS w/o VGD, the curves of high gradient variance (blue and yellow) remain relatively high as the densification strategy progresses. For 3D-GS without VGD, the high gradient variance curves (blue and yellow) remain relatively elevated as the densification strategy progresses. In contrast, for 3D-GS with VGD, the corresponding high gradient variance curves decrease rapidly, while the low gradient variance curves (green and red) show an increase. These results show that our suggested method of densification successfully decreases color discrepancies among pixels in Gaussians.}
\label{fig3}
\end{figure}

To validate our hypothesis, we conduct experiments by adjusting the CUDA differential rasterization pipeline. During training, we compute the mean and variance of the color gradients for Gaussians to track how the mean and variance of pixel gradients change towards a Gaussian distribution. Results are shown in Figure \ref{fig3}. The findings indicate that many Gaussians still show a wide range of color gradients, even after completing the final stage of training, suggesting that the rendered pixels are not fitting well. 

This analysis highlights the difficulties presented by intricate textures, as the inconsistency in pixel gradients makes it difficult to densify and fit effectively. This observation also intuitively matches the characteristics of areas that have not been fully reconstructed as perceived by humans. Although current densification methods prioritize improving gradients in high-frequency or inadequately fitted regions to facilitate the satisfaction of densification criteria, there is still a lack of statistical approach to tackling this problem.

\subsection{Variance-Guided Densification}
In 3DGS, conventional methods for densification involve using positional gradient magnitude to decide the timing for Gaussian splitting and cloning. Yet, this method may not work well in regions with intricate patterns, as the orientations of gradients could differ greatly.

We propose a Variance-Guided Densification (VGD) approach to overcome this restriction. Our method takes into account the color gradient variances for pixels rendered by each Gaussian point in the backpropagation process. This method works well in pinpointing areas that are not adequately represented by a single Gaussian point due to their variety of textures or colors. We use an iterative method within the rasterizer to calculate the mean and variance of pixel gradients for each Gaussian point during backpropagation. The following equations outline the process:
\begin{equation}
   \hat{\mu}_{n+1}=\hat{\mu}_{n}+\beta_{n+1}(g_{n+1}-\hat{\mu}_{n})
\end{equation}
\begin{equation}
   \hat{\sigma}^{2}_{n+1}=(1-\beta_n)\hat{\sigma}^2_{n}+\beta_{n+1}(g_{n+1}-\hat{\mu}_{n})^2
\end{equation}
Here, $n$ denotes the index of the pixel rendered by the Gaussian, $\beta_n = 1/n$, and $g_n$ represents the backpropagated gradient of the $n$-th pixel. We perform these calculations separately for each RGB channel to capture color-specific variations. When updating the accumulated status, we calculate the total variance by adding up the variances from the RGB channels to accurately gauge the Gaussian's overall deviation. This method gives a strong indication of how well the Gaussian model can accurately represent the local color distribution.

We adjust this strategy to work with adaptive densification by rescaling the value to fit the original threshold's magnitude. The final determinate criterium is the scaled maximum of the original average norm with our average variant:
\begin{equation}
    \gamma\bar{D} +\bar{g}_\mathrm {norm}>\tau_\mathrm{th}
\end{equation}
\begin{equation}
    \bar{D} =\frac{\sum_{k=1}^{M}\hat{\sigma}^{2}}M
\end{equation}
where $M$ is the number of views rendered in densification interval. Our approach utilizes the current CUDA implementation of the rasterizer's backpropagation process to effectively calculate the mean and variance of pixel gradients simultaneously for each Gaussian point, seamlessly integrating the computations with minimal computational burden.

\begin{figure*}[h]
\centering
\includegraphics[width=0.99\textwidth]{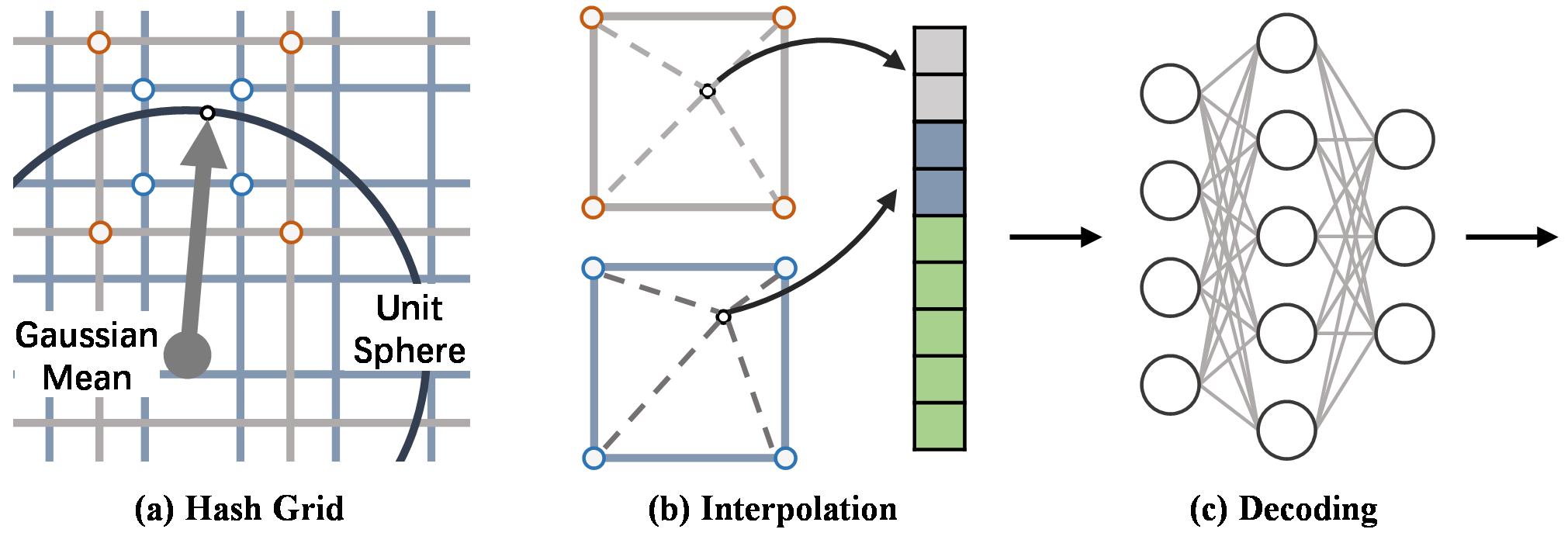} 
\caption{\textbf{Illustration of Lighting Hash Encoder projected into 2D.} (a) We index the hash grid by Distant Gaussians are normalized and projected onto a unit sphere. The vertices of the hash grid cells intersected by this unit sphere represent the hash grid parameters that will be optimized during training. This allows for the effective encoding of view-dependent lighting information.}
\label{hash-pipeline}
\end{figure*}

\subsection{Lighting Hash Encoder}
3DGS utilizes spherical harmonics to depict color variations in various view angles, capturing photometric inconsistencies caused by specular reflection, diffusion, and ambient light in the scene. Scaffold-GS, on the other hand, directly connects implicit feature embeddings with viewing directions, and an MLP decodes the color of the Gaussian sphere from different angles. 

However, in certain cases where lighting is complicated, just inputting the view direction may not effectively encourage the model to understand how color varies across different light sources. When using vanilla Gaussian-splatting, spherical harmonics are unable to capture the complex lighting patterns. In Scaffold-GS, the size of the anchor embedding restricts its ability to accurately represent varying light conditions from different viewpoints.

The hash grid efficiently encodes spatial information by mapping complex spatial data into a compact hash table, allowing for efficient storage and quick querying. This method keeps the amount of data high but decreases the amount of storage space and computational burden. Our suggestion is to utilize a hash grid for encoding the lighting details of the scene. 

To calculate the 3-dimensional view direction vector for each anchor, we normalize the difference between the positions of the anchors and the camera. The vectors can take on any values that fall within the range of [-$1,1$], and collectively they form a unit sphere. This also defines the bounding box of the hash grid. Next, we identify the adjacent voxels and apply a hash function to convert the vectors into hash indices. The hash function can be defined in the following way:
\begin{equation}
    h(\mathbf{x}) = \left( \left\lfloor \frac{x}{s_x} \right\rfloor, \left\lfloor \frac{y}{s_y} \right\rfloor, \left\lfloor \frac{z}{s_z} \right\rfloor \right) \bmod M
\end{equation}
where $s_x, $ $s_y, $ and $s_z$ are the minimal resolutions of current level and $M$ is the size of the hash table. These indices can quickly locate relevant cells in the hash grid. Trilinear interpolation is utilized to determine the precise values at the vector positions, blending the lighting values seamlessly according to the specific location of the query position within the grid. Picture yourself blending colors from various paint containers to achieve a seamless transition.

Although the hash grid is effective for encoding light conditions, it tends to overfit the training views. Overfitting can result in inadequate generalization when used on test samples, which can reduce the model's reliability and precision. We add random noise to the view direction vector to ensure the generalization of our model. This helps to achieve a more seamless and consistent color response within the actual viewing directions, thus mitigating the risk of overfitting and promoting better performance on test data. 

\section{Experiments}
\subsection{Experimental Setup}
\textbf{Dataset} We evaluate our method on a diverse range of scenes from three primary datasets: Mip-NeRF 360 \citep{mipnerf360}, Tanks \& Temples \citep{tnt}. and DeepBlending \citep{deepblending}. This comprehensive selection includes both real-world and synthetic environments, encompassing indoor and outdoor scenes with varying levels of complexity. The MipNeRF-360 dataset provides seven real-world scenes with challenging view distributions, while NeRF Synthetic offers controlled synthetic scenes ideal for baseline comparisons. Tanks \& Temples contributes large-scale real-world scenes with intricate geometries, and DeepBlending adds diversity with its unique capture methodology. These varied datasets allow us to test the generalization ability of our method across different scene types and capture conditions. Following the protocol established in Mip-NeRF360 \citep{mipnerf360}, we designate every eighth image as part of the test set, utilizing the remaining images for training, ensuring a consistent and fair evaluation across all methods.

\begin{table*}[pos=ht,width=1\textwidth,cols=10]
\small
\centering
\renewcommand{\arraystretch}{1.2}
\caption{\textbf{Comparison with different novel view synthesis methods across three public datasets (Mip-NeRF 360\citep{mipnerf360}, Tanks\&Temples\citep{tnt}, and Deep Blending\citep{deepblending}). Results are presented in SSIM \citep{wang2004image}, PSNR, and LPIPS \citep{zhang2018unreasonable}.}}
    \begin{tabularx}{\tblwidth}{lXXXXXXXXX}
    \toprule
    \multirow{2}{*}{Method} & \multicolumn{3}{c}{Mip-NeRF360} & \multicolumn{3}{c}{Tanks\&Temples} & \multicolumn{3}{c}{Deep Blending} \\
     & SSIM$\uparrow$ & PSNR$\uparrow$ & LPIPS$\downarrow$ & SSIM$\uparrow$ & PSNR$\uparrow$ & LPIPS$\downarrow$ & SSIM$\uparrow$ & PSNR$\uparrow$ & LPIPS$\downarrow$ \\
    \midrule
    INGP-Base &0.671  & 25.30   & 0.371   & 0.723  & 21.72  & 0.330  & 0.797  & 23.62   & 0.423 \\ 
    INGP-Big  &0.699  & 25.59   & 0.331   & 0.745  & 21.92  & 0.305  & 0.817  & 24.96   & 0.390 \\ 
    M-NeRF 360&0.792  & 27.69   & 0.237   & 0.759  & 22.22  & 0.257  & 0.901  & 29.40   & 0.245 \\ 
    3DGS        &0.870  & 29.07   & 0.184   & 0.841  & 23.14  & 0.183  & 0.881  & 28.92   & 0.287 \\
    Scaffold  &0.848  & 28.84   & 0.220   & \textbf{0.853}  & 23.96  & 0.177  & 0.906  & 30.21   & 0.254 \\
    Ours      & \textbf{0.876} & \textbf{29.52}   & \textbf{0.171}  
              & 0.850          & \textbf{24.03}   & \textbf{0.176}  
              & \textbf{0.912}          & \textbf{30.40}   & \textbf{0.230} \\
    \bottomrule
    \end{tabularx}
\label{table1}
\end{table*}


\begin{figure*}[t]
\centering
\includegraphics[width=0.99\textwidth]{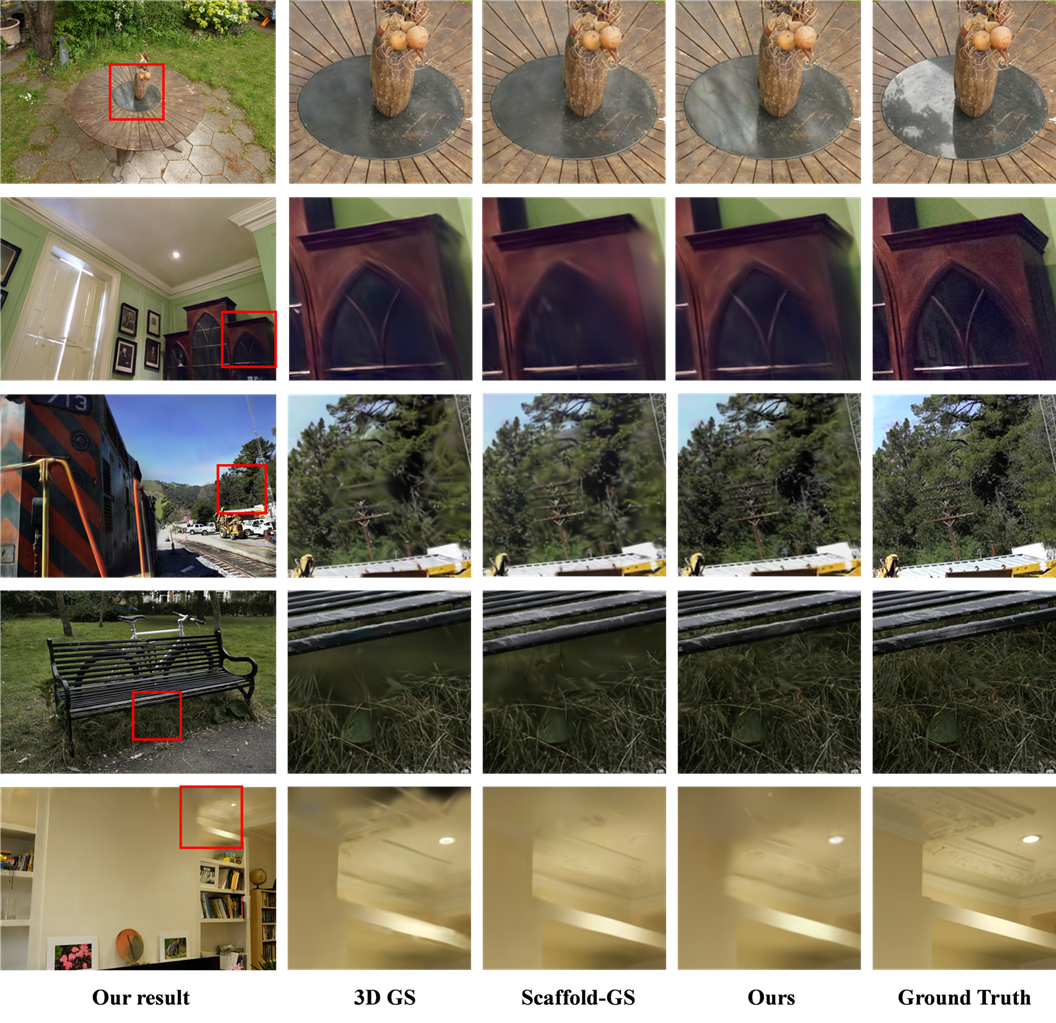}
\caption{\textbf{Comparison on multiple datasets.} We conducted extensive experiments on diverse datasets including indoor, outdoor, and object-centric datasets. Results show that our method outperforms the predecessor, providing better rendering effects for
texture details.}
\label{compare1}
\end{figure*}
\noindent\textbf{Metrics} To quantitatively assess the performance of our novel view synthesis method, we employ three complementary metrics: SSIM \citep{wang2004image}, PSNR, and LPIPS \citep{zhang2018unreasonable}. SSIM evaluates the structural and perceptual similarity between the synthesized views and ground truth, capturing local patterns of pixel intensities. PSNR provides a standard measure of reconstruction quality, particularly sensitive to per-pixel differences. LPIPS offers a perceptual metric that aligns well with human visual perception, evaluating differences in feature space rather than pixel space. Together, these metrics provide a comprehensive evaluation of our method's ability to generate high-quality, perceptually accurate novel views.

\noindent\textbf{Implementation} We conducted all experiments on an RTX 3090 GPU, maintaining consistent hyperparameters across all scenes to demonstrate the robustness and generalizability of our method. Our experimental setup encompasses both quantitative and qualitative comparisons with existing state-of-the-art novel view synthesis methods, as well as detailed ablation studies on our key technical components. These studies aim to validate the effectiveness of each component in our proposed Metamon-GS. For modeling lighting, we use an 8-level hash grid with a base resolution of 8, the maximum hash map size is 19. The coefficient $\gamma$ we adopted is $2^{11}$. We set $\tau_\mathrm {th}$ to 0.0004 following our baseline model.

\subsection{Results}
In our comparative experiments, we evaluated our approach against several leading methods across 11 real-world scenes sourced from three aforementioned datasets, providing a comprehensive benchmark of our method's capabilities in diverse real-world scenarios. We compare our model with 4 predecessors, including Mip-NeRF 360, Instant-NGP \citep{ingp}, 3DGS \citep{3dgs}, and Scaffold-GS \citep{scaffold}. As shown in Table \ref{table1}, our method outperforms these predecessor approaches. The results demonstrate that our method surpasses the state-of-the-art model by 0.45 dB in PSNR on Mip-NeRF360 dataset. 

Rendering results presented in Figure \ref{compare1} clearly demonstrate that our model exhibits remarkable perception capabilities for environmental light changes, such as the reflections of the environment on the surfaces of table and cabinet, as well as for complex textures, like the sharp edges on leaves and grassland. Our model shows a strong capability in modeling intricate details, and its scene reconstruction ability has significantly surpassed that of the baseline and other advanced models.


\begin{table}[width=1\linewidth,cols=4,pos=h]
\caption{\textbf{Ablation studies on our proposed light encoding and densification strategy.} Variance-guided densification adds sufficient Gaussians to better reconstruct the scene and Hash Grid Encoded Lighting enables better color representation considering different view directions.}\label{table2}
\renewcommand{\arraystretch}{1.2}
\begin{tabularx}{\tblwidth}{lXXX}
\toprule
    Method    & SSIM$\uparrow$ & PSNR$\uparrow$  & LPIPS$\downarrow$ \\ 
\midrule
Base      &      0.848 &  28.84         &   0.220          \\ 
Base+LHE     &      0.870 &   29.34       &   0.187        \\ 
Base+LHE+VGD &      \textbf{0.876} &   \textbf{29.52}       &   \textbf{0.171}   \\
\bottomrule
\end{tabularx}
\end{table}


\subsection{Ablation Study}
We conducted an ablation study to evaluate the effectiveness of the separate parts of our proposed method on the Mip-NeRF 360 \citep{mipnerf360} dataset. Mip-NeRF 360 is a widely-used dataset in the field of 3D reconstruction, featuring diverse scenes and complex textures, which provides a challenging environment to test the robustness and effectiveness of our method. 

We evaluated the variance-guided densification strategy and found that it effectively facilitated the Gaussians to densify in areas with complex textures. In complex-textured regions, the variance of pixel color gradients is typically high. By analyzing this variance, our strategy can precisely identify these areas and allocate more Gaussians, enhancing the overall fitting to the scene. This improvement also led to an increase in rendering quality.

Results are presented in Table \ref{table2}. In terms of evaluation metrics, the proposed LHE achieved an improvement in PSNR by 0.18 dB, and a reduction in LPIPS by 0.016. 
These improvements can be attributed to the unique design of LHE. By using a hash grid to encode lighting information and considering lighting conditions as a global attribute with directional information, LHE can better capture the complex lighting variations in the scene. This enables more accurate color representation under different view directions, thus contributing to the overall improvement in rendering quality.

Overall, our ablation study demonstrates the effectiveness of both the variance-guided densification strategy and the Lighting Hash Encoder in enhancing the performance of our proposed method on the challenging Mip-NeRF 360 dataset.

\section{Conclusion}
We have introduced a new method for identifying Gaussians that need to be densified, which is based on the variance of color gradients in pixels generated. This method corresponds to how humans perceive blurriness in areas that are not well-fitted. Furthermore, a view-dependent hash grid feature is implemented to substitute the view direction vector input of the color MLP, reducing the uncertainty in modeling intricate lighting for Gaussian anchors.The results of the experiment show that our new method is effective, outperforming other methods in novel view synthesis tasks. Our model generates high-quality images with enhanced detail and fewer defects when compared to current methods.

However, our approach exhibits some limitations. The view-dependent modeling remains sensitive to extreme viewpoint variations beyond the training distribution, occasionally causing color inconsistency in under-observed regions. Additionally, the densification criteria require careful parameter tuning for scenes with complex multi-scale structures, as aggressive densification could accidentally lead to unnecessary Gaussian growth in textureless regions.

Future work will focus on two directions: first, developing a geometry-aware optimization to improve robustness against extreme perspective variations, and second, designing self-adaptive densification criteria capable of automatically adjusting to local structural complexity. These advancements will further strengthen the practicality of Gaussian splatting in real-world 3D Reconstruction.


\printcredits

\bibliographystyle{cas-model2-names}

\bibliography{cas-refs}

\end{document}